\documentclass[envcountsame]{article}

\usepackage[utf8]{inputenc}
\usepackage[english]{babel}
\usepackage[hidelinks]{hyperref}
\usepackage[table]{xcolor}
\usepackage{graphicx}
\usepackage{comment}
\usepackage{amsthm}
\usepackage{amsmath}
\usepackage{todonotes}
\usepackage{empheq}

\usepackage[sc]{mathpazo}

\newcommand{\PbOpt}[3]{%
\begin{center}
  \begin{tabular}{|l|}%
  \hline
    \begin{minipage}[c]{.95\textwidth}
    \smallskip%
      \par\noindent%
      \textsc{#1:}
      \par\noindent%
      $\bullet$
      \textbf{\textsf{Input}}: #2%
      \par\noindent%
      $\bullet$
      \textbf{\textsf{Output}}: #3%
      \smallskip%
      \par\noindent%
    \end{minipage}
    \\\hline
  \end{tabular}%
\end{center}
}%

\newtheorem{funfact}{Fun Fact}

\title{The shortest way to visit all metro lines in a city}
\date{}

\author{Florian Sikora
\footnote{\noindent Universit\'{e} Paris-Dauphine, PSL Research University, CNRS, LAMSADE, Paris, France.  \tt{florian.sikora@dauphine.fr}}
}

\begin{document}
\maketitle

\begin{abstract}
What if $\{$a tourist, a train addict, Dr. Sheldon Cooper, somebody who likes to waste time$\}$ wants to visit all metro lines or carriages in a given network in a minimum number of steps?
We study this problem with an application to the metro network of Paris and Tokyo, proposing optimal solutions thanks to mathematical programming tools.
Quite surprisingly, it appears that you can visit all 16 Parisian metro lines in only 26 steps (we denote by a step the act of taking the metro from one station to an adjacent one). 
Perhaps even more surprisingly, adding the 5 RER lines to these 16 lines does not increase the size of the best solution. 
It is also possible to visit the 13 lines of (the dense network of) Tokyo with only 15 steps.
\end{abstract}

\section{Introduction and History}

During a lunch in the early 2017, a colleague researcher, Jérôme Lang, asked what would be the shortest journey in the Parisian metro which uses all metro lines at least once (and promised to take the tour once found).
It appears that it is close to a known challenge for the New-York subway, denoted ``subway challenge'' (or ``Rapid Transit Challenge'', or ``Ultimate Ride'', or also ``Tube Challenge'' in London), where the time to complete the challenge must be minimized.
Three variants of the challenge were defined by the Amateur New York Subway Riding Committee (ANYSRC)~\cite{ANYSRC}, in decreasing order of time needed to complete it:
\begin{itemize}
\item Class A: ``Covering all lines'': the contestant must use every portion of the metro, between each any connected station\footnote{Note that there is an error in the corresponding Wikipedia article (at least in February 2018~\cite{WikipediaSubway}), where it is stated that the contestant must ``traverse every line, but not necessarily the entire line'', but the rules of the ANYSRC are different and require to visit every portion of each line.}.
\item Class B: ``Touching all stations'': the contestant must stop at each station.
\item Class C: ``Passing all stations'': the contestant must pass by every station, but can use express trains which are not stopping at some stations.
\end{itemize}

The class B challenge is recognized by the Guinness Book of Records~\cite{GuinnessWorldRecords2015,GuinnessWorldRecords2017}. 
The current record for the Class B challenge in Paris seems to be by Adham Fisher in 13h 37$'$ 54$''$, to go through the 302 stations, done in 2011~\cite{Mole2011,HuffPost}.
No details are given on how he decided the order of the stations.
Some students of the Ecole Normale Sup\'erieure did a similar journey but without considering stations outside Paris~\cite{Madore2004}.
They used a computer to find a ``not too bad'' journey and they went through 242 stations in 12h $22'$ 03$''$.
A French YouTuber also tried to visit all lines of the Parisian metro~\cite{Tatou2015}. 
He spent 15h 13$'$ but he seems to have visited all lines from 1 to 14 in order, without any optimization of the journey.

Some theoretical work also has been done around these classes of challenges.
In his Ph.D. Thesis~\cite[Chapter 6]{Welz2014}, Welz studied the subway challenge of class A and class B with formulations and applications to the Berlin network in a model taking into account uniform changing and waiting times.
Beckenbach et al. consider the correct transfer time for each station as well as the waiting time for each time of the day and each train schedule in order to have a more accurate total time! 
They transformed their problem into a Travelling Salesman Problem (TSP) in order to use an efficient TSP solver.
They applied their model to the S-Bahn network of Berlin.

\medskip

The question of finding the shortest journey which uses all metro lines is therefore not a class already defined so far.
However, it appears that it is a known problem in Operational Research.
Let us first give some light graph definitions in order to formalize our problem.
A metro network can be represented as a directed (multi)-graph, where each vertex is a metro station and where there is an arc colored by the metro line $\ell$ between two stations if the line $\ell$ link the two corresponding stations.
A walk in a graph is a sequence of vertices such that every consecutive pair of vertices is joined by an arc.
A vertex in a walk can be used more than once.
Our problem is thus to find the shortest walk between two vertices of the graph using at least once each color of the arcs.
This is known under the name of \textsc{Generalized Directed Rural Postman} problem, as defined by Drexl~\cite{Drexl2014} (note that the original problem aims at finding a tour in graph instead of a walk).

\PbOpt{Generalized Directed Rural Postman (GDRP)}{A set of colors $C$, a (directed multi)-graph $G=(V,A)$, an arc-coloring function $c : A \to C$.}{The shortest walk $S$ between two nodes of $V$ s.t. 
$\cup_{a \in S} c(a) = C$.
}

Drexl gave a formulation of GDRP as well as some exact algorithm to solve it~\cite{Drexl2014}.
{\'{A}}vila et al. gave additional formulation and branch-and-cut algorithms which are faster~\cite{Avila2016}.
However, both formulations seems to use exponential number of constraints.

GDRP is related to a broader class of problems called ``Arc Routing Problems''~\cite{Corberan2013}. 
The goal in these problems is to select the best path but with a focus on the properties of the arcs (instead of the nodes).
A famous problem in this class is the \textsc{Chinese Postman} problem where one must find a minimum-cost tour which visit all the arcs of a given graph at least once. 
This problem can be solved in polynomial-time as proven by Edmonds and Johnson~\cite{Edmonds1973}.
Another famous problem is the \textsc{Rural Postman} problem, a generalization of the previous problem, where only a subset of the arcs must be visited. 
This problem is however NP-hard (i.e. not solvable in polynomial-time, unless P=NP)~\cite{Lenstra1976}.

To the best of our knowledge, nobody applied GDRP to a metro network.
In the following, we will present our formulation as an ILP and give some experimental results for the metro network in Paris and in Tokyo.


\section{Formulation}

In the decision variant of GDRP, an additional integer $k$ is given and the question is to find a walk of length at most $k$.
Unsurprisingly, GDRP is NP-hard
~\cite{Drexl2014}.

But still, we need a solution to this intriguing problem for a metro network, in order to do the tour!
We thus decided to model this problem using Integer Linear Programming (ILP). 
Such formulation permits the use of efficient tools to solve it and to obtain optimal solution in reasonable time.

\begin{align}
	&minimize \sum_{(u,v,\ell) \in A} x_{u,v,\ell}
    \label{obj}
\end{align} 
Subject to:
\begin{empheq}[left=\empheqlbrace]{align} 
&	\sum_{(u,v,\ell) \in A  } x_{u,v,\ell} = \sum_{(v,w,\ell) \in A  } x_{v,w,\ell}, & \forall& v \in V \setminus \{s,t\}
    \label{constr:degree}\\
 &   \sum_{(s,v,\ell) \in A  } x_{s,v,\ell} = \sum_{(u,t,\ell) \in A  } x_{u,t,\ell} = 1
        \label{constr:degree_st}\\
 & \sum_{(u,v,\ell) \in A  } x_{u,v,\ell} \geq 1,& \forall& \ell \in C
            \label{constr:lines}\\
 & 	|V|\cdot x_{u,v,\ell} \geq f_{u,v,\ell}, &\forall &(u,v,\ell) \in A
                \label{constr:flow1}\\
& \sum_{(u,v,\ell) \in A} f_{u,v,\ell}  - \sum_{(v,w,\ell) \in A} f_{v,w,\ell} \geq y_v    , &\forall& v\in V \setminus \{s\}
                    \label{constr:flow2}\\
&	y_v - \sum_{(u,v,\ell) \in A} x_{u,v,\ell} - \sum_{(v,w,\ell) \in A} x_{v,w,\ell} \geq 0, &\forall& v \in V
    \label{constr:visited}\\
&	x_{u,v,\ell} \in \{0,1\}, f_{u,v,\ell} \in \mathbb{N}, &\forall& (u,v,\ell) \in A
        \label{vars1}\\
&	y_{v} \in \mathbb{N},&\forall& v \in V
    \label{vars2}
\end{empheq} 

Our formulation is quite easy. 
We will consider binary variables $x_{u,v,\ell}$ for every arc from $u$ to $v$ on the line $\ell$, which are set to 1 if the corresponding arc is taken in the solution, 0 otherwise.
We want to minimize the sum of these variables.
We add a fake vertex source (denoted $s$) with an arc to every station and a fake vertex target (denoted $t$) with an arc from every station.
We look for a shortest walk from the source to the target, that is, for every station (except the fake nodes), the sum of the outgoing arcs must be equal to the sum of the in-going ones (Constraints~\ref{constr:degree} and \ref{constr:degree_st}).
We want to visit all lines: therefore, for every line, the sum of the arcs corresponding to this line must be at least 1 (Constraint~\ref{constr:lines}).
To enforce the solution to be connected, we add an auxiliary flow (Constraints~\ref{constr:flow1} and \ref{constr:flow2}) (as explained for example in~\cite{Cerny2015}).
To do so, we need variables $y_v$, which are positive if station $v$ is in the solution (with Constraint~\ref{constr:visited}) and variables $f_{u,v,\ell}$, which are non-negative if the arc is visited (Constraint~\ref{constr:flow1}). 
Only the source generate flow, while every other vertex in the solution loses some flow.
Indeed, if $y_v$ is positive, then Constraint~\ref{constr:flow2} implies that the sum of the incoming flow to $v$ is greater than the outgoing flow from $v$.

For different experiments, we need sometimes to modify a bit this program.
If one wants to avoid the use of the same station multiple time in the solution, we can bound the maximum value of variables $y_v$ in the formulation (Constraint~\ref{vars2} is then $y_{v} \leq 2$).
If one wants to avoid a solution which uses a line which was already taken previously in the solution, we need new binary variables and new constraints.
The new binary variables tell if two consecutive edges are together in the solution:  $z_{u,v,w,\ell_1, \ell_2} \in \{0,1\}, \forall (u,v,\ell_1), (v,w,\ell_2) \in A$ with the constraints $x_{u,v,\ell_1} + x_{v,w,\ell_2} \leq z_{u,v,w,\ell_1,\ell_2} + 1$ (therefore $z_{u,v,w,\ell_1,\ell_2}$ is true iff both $x_{u,v,\ell_1}$ and $x_{v,w,\ell_2}$ are true). 
Then, we can add the constraints, for all $\ell \in C$, $\sum_{ (u,v,\ell_1), (v,w,\ell_2) \in A \text{ s.t. } (\ell = \ell_1 \text{ or } \ell = \ell_2)} z_{u,v,w,\ell_1,\ell_2} = 2$, avoiding taking more than once a line $\ell$.
If one wants to count the number of optimal solutions, we will solve multiple times the model.
For each solved model, let $X = \{ x_{u,v,\ell} | x_{u,v,\ell} = 1 \}$, i.e. $X$ represents the solution of the journey. 
We add $\sum_{x \in X} x < |X|$ as a new constraint, and solve the model another time. 
This new constraint avoid the possibility to have the same solution again. 
We stop this process when the next model have a strictly worse solution.

\section{Experimental results}

We build a graph with an edge between two stations if there is a line connecting them, or if it is possible to go from one station to the other without exiting the network (this is drawn on the metro maps).
It is therefore not possible to walk outside the metro network to go to the next step.
A light pre-processing is also applied to the graph. 
Indeed, we can safely (recursively) remove from the graph all vertices which are the terminus of a line and not connected to other lines of the network (it is vertices with only one neighbor). 

We used CPLEX 12.5~\cite{ILOG2012} to solve the linear program described above. 

\subsection{Optimal journeys in Paris}

The Parisian network is made of 16 lines of metro (numbered from 1 to 14, including 3bis and 7bis) and 5 RER lines (named A to E).
The data for Paris is collected from a teaching assignment~\cite{Chassignet2004} and the missing data is completed manually.

We first compute the number of steps that can be used to visit all the metro lines at least once in Paris.

\begin{funfact}\label{fact:metro} It is possible to visit the 16 lines of the Parisian metro network with 26 steps.
\end{funfact}

\begin{figure}
\centering

\includegraphics[width=\linewidth]{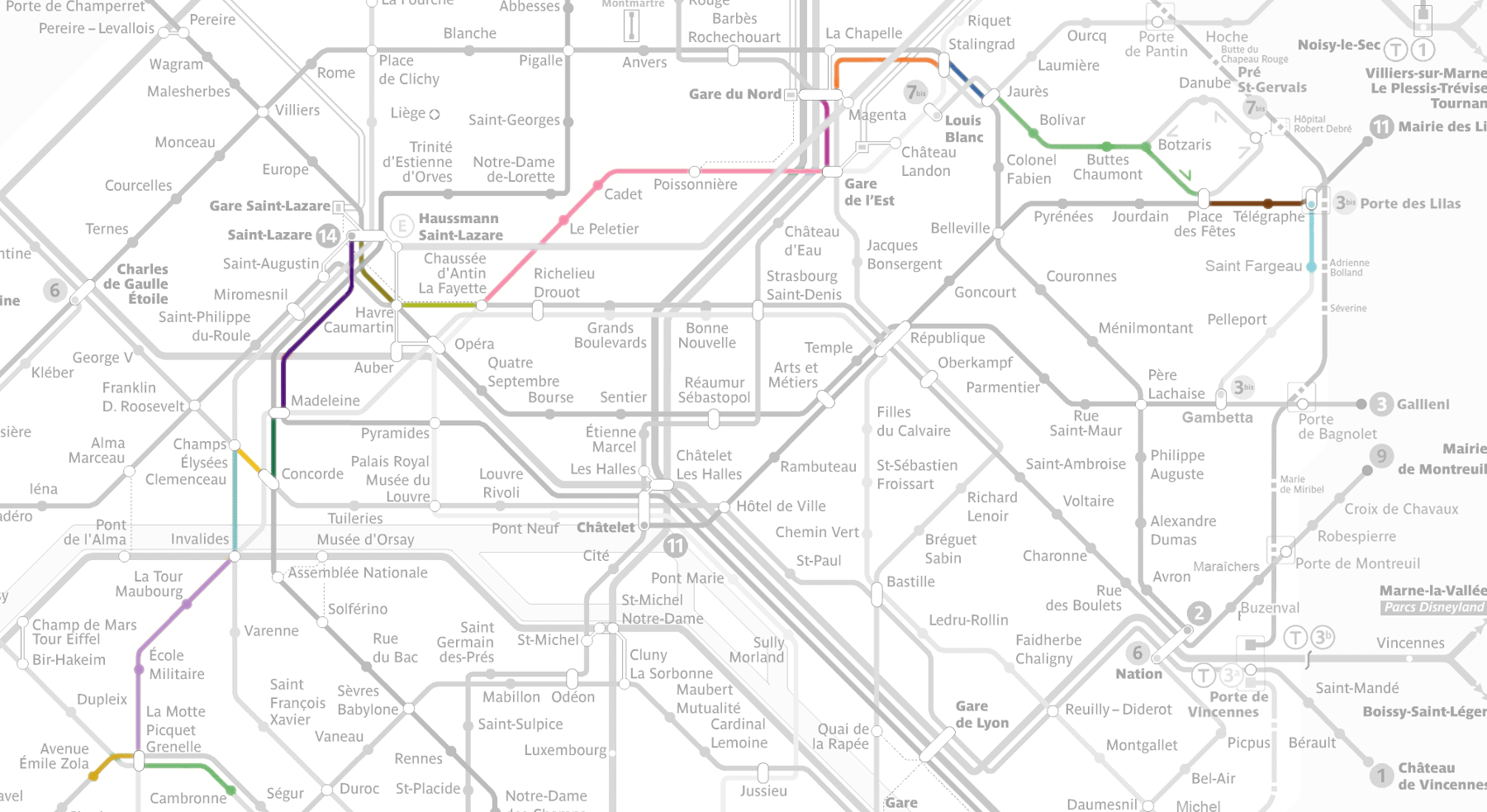}
\caption{Map for an optimal journey including all 16 metro lines, from Cambronne to Saint Fargeau.}
\label{fig:metro1}
\end{figure}

The details of one such journey, from ``Cambronne'' to ``Saint-Fargeau'' is given in Table~\ref{tab:metro1} and drawn in Figure~\ref{fig:metro1}.
There are many other solutions with the same number of steps (we could not compute the exact number since the computation was not finished after three weeks). 
For a simple example, one can also starts from ``Avenue Emile Zola'' and do step 3, ``La Motte-Piquet-Grenelle'' to ``Cambronne'' by line 6 and step 1, and then continue with the same journey from step 4.


\begin{table}
\centering
\resizebox{0.9\columnwidth}{!}{
\begin{tabular}{llll}
Step & Departure & Arrival & Line \\\hline
\rowcolor[HTML]{6eca97}
1&Cambronne&La Motte-Picquet - Grenelle&6 \\ 
\rowcolor[HTML]{c9910d}
2&La Motte-Picquet - Grenelle&Avenue Emile Zola&10 \\ 
\rowcolor[HTML]{c9910d}
3 & Avenue Emile Zola & La Motte-Picquet - Grenelle & 10  \\ 
\rowcolor[RGB]{225, 155, 223}
4&La Motte-Picquet - Grenelle&Ecole Militaire&8 \\ 
\rowcolor[RGB]{225, 155, 223}
5&Ecole Militaire&La Tour-Maubourg&8 \\ 
\rowcolor[RGB]{225, 155, 223}
6&La Tour-Maubourg&Invalides&8 \\ 
\rowcolor[RGB]{ 	110, 196, 232}
7&Invalides&Champs Elysees - Clemenceau&13 \\ 
\rowcolor[RGB]{255, 205, 0}
8&Champs Elysees - Clemenceau&Concorde&1 \\ 
\rowcolor[HTML]{007852}
9&Concorde&Madeleine&12 \\ 
\rowcolor[HTML]{62259d}
\textcolor{white}{10}&\textcolor{white}{Madeleine}&\textcolor{white}{Saint-Lazare}&\textcolor{white}{14} \\ 
\rowcolor[HTML]{837902}
11&Saint-Lazare&Havre - Caumartin&3 \\ 
\rowcolor[HTML]{b6bd00}
12&Havre - Caumartin&Chaussee d'Antin - La Fayette&9 \\ 
\rowcolor[RGB]{ 	250, 154, 186}
13&Chaussee d'Antin - La Fayette&Le Peletier&7 \\ 
\rowcolor[RGB]{ 	250, 154, 186}
14&Le Peletier&Cadet&7 \\ 
\rowcolor[RGB]{ 	250, 154, 186}
15&Cadet&Poissonniere&7 \\ 
\rowcolor[RGB]{ 	250, 154, 186}
16&Poissonniere&Gare de l'Est&7 \\ 
\rowcolor[RGB]{207, 0, 158}
17&Gare de l'Est&Gare du Nord&4 \\ 
\rowcolor[RGB]{ 	255, 126, 46}
18&Gare du Nord&Stalingrad&5 \\ 
\rowcolor[RGB]{0, 60, 166}
\textcolor{white}{19}&\textcolor{white}{Stalingrad}&\textcolor{white}{Jaures}&\textcolor{white}{2} \\ 
\rowcolor[HTML]{6eca97}
20&Jaures&Bolivar&7bis \\ 
\rowcolor[HTML]{6eca97}
21&Bolivar&Buttes Chaumont&7bis \\ 
\rowcolor[HTML]{6eca97}
22&Buttes Chaumont&Botzaris&7bis \\ 
\rowcolor[HTML]{6eca97}
23&Botzaris&Place des Fetes&7bis \\ 
\rowcolor[HTML]{704b1c}
\textcolor{white}{24}&\textcolor{white}{Place des Fetes}&\textcolor{white}{Telegraphe}&\textcolor{white}{11} \\ 
\rowcolor[HTML]{704b1c}
\textcolor{white}{25}&\textcolor{white}{Telegraphe}&\textcolor{white}{Porte des Lilas}&\textcolor{white}{11} \\ 
\rowcolor[HTML]{6ec4e8}
26&Porte des Lilas&Saint-Fargeau&3bis
\end{tabular} 
}
\caption{An optimal journey with 26 steps visiting the 16 metro lines.}
\label{tab:metro1}
\end{table}

\begin{funfact}
The solution given in Fun Fact~\ref{fact:metro} cannot be traveled on the other direction due to the portion using line 7bis which cannot be used in both directions.
\end{funfact}

One can now ask what would be the best solution if, in addition to the 16 metro lines, we add the 5 RER lines. 
RER lines go far outside Paris but all have portion inside Paris with connections to the metro lines.
Quite surprisingly, this can still be done in 26 steps! 
The use of RER portion permits to quickly ``jump'' from two different parts of Paris and avoid long portions of the same line as used in the metro only solution (especially the 4 steps with line 7).

\begin{funfact}
\label{fact:rer} It is possible to visit the 16 lines of the Parisian metro network and the 5 RER lines with 26 steps, and there are 20 ways to do it.
\end{funfact}

\begin{figure}
\centering
\includegraphics[width=\linewidth]{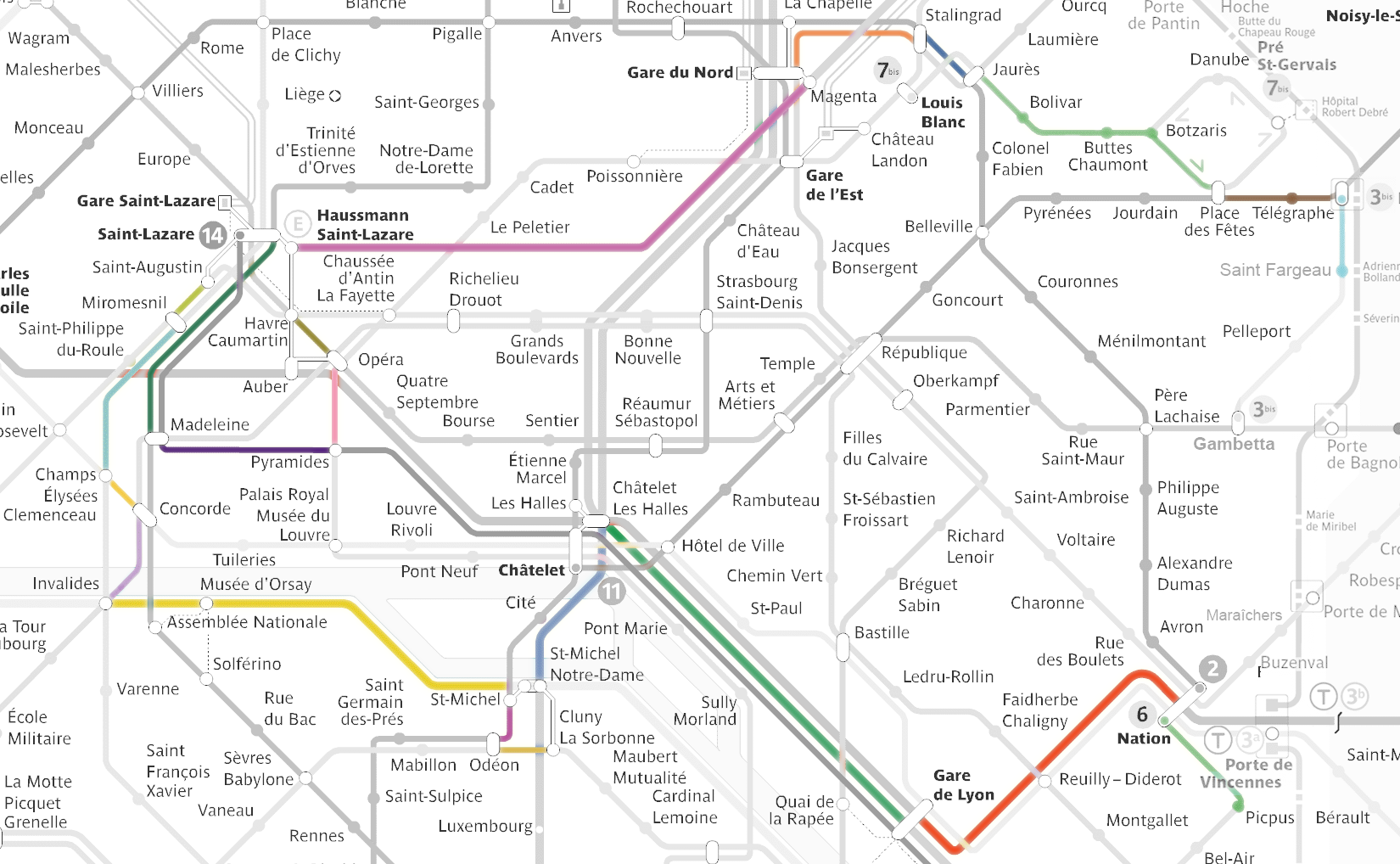}
\caption{Map for an optimal journey including all 16 metro lines and 5 RER lines, from Picpus to Saint Fargeau.}
\label{fig:rer}
\end{figure}

The details of one of these journeys using all metro and RER lines, from ``Picpus'' to ``Saint-Fargeau'', is given in Table~\ref{table:rer} and drawn in Figure~\ref{fig:rer}.
Some of these 20 equivalent journeys can be traveled in both direction and remain optimal because they use the part of the line 7bis which is undirected (contrary to the one shown in Table~\ref{table:rer}).


One could now legitimately ask if allowing RER portions without asking to visit all RER lines would shorten the ``visit all metro lines'' journey.
Unsurprisingly, this is the case, with 2 step less.
However, this can be seen as a sort of cheating.

\begin{funfact}
If one can use the RER (at the same cost as a metro step), it is possible to visit the 16 metro lines with only 24 steps.
\end{funfact}

A meticulous reader will observe that both solutions of Fun Fact~\ref{fact:metro} and Fun Fact~\ref{fact:rer} are visiting more than once  some stations. 
More saddening, the journey for the metro given by Fun Fact~\ref{fact:metro} uses a ``back and forth'' to visit line 10. 
A tourist may not desire to go two times on the same station!
What would be the best journey if one does not allow such steps?
In other terms, what is the size of the solution if we request a path instead of a walk?
Not surprisingly, it increases a bit the number of steps of the optimal solution, and quite dramatically for the version where RER lines must also be visited.

\begin{funfact}\label{fact:nostation}
If it is not allowed to use more than once a same station, it is possible to visit all metro lines in 27 steps (instead of 26 steps) and there are 5 ways to do it. 
Also, it is possible to visit all metro lines and all RER lines in 29 steps (instead of 26 steps) and there are 9 ways to do it.
\end{funfact}

\begin{figure}
\centering
\includegraphics[width=\linewidth]{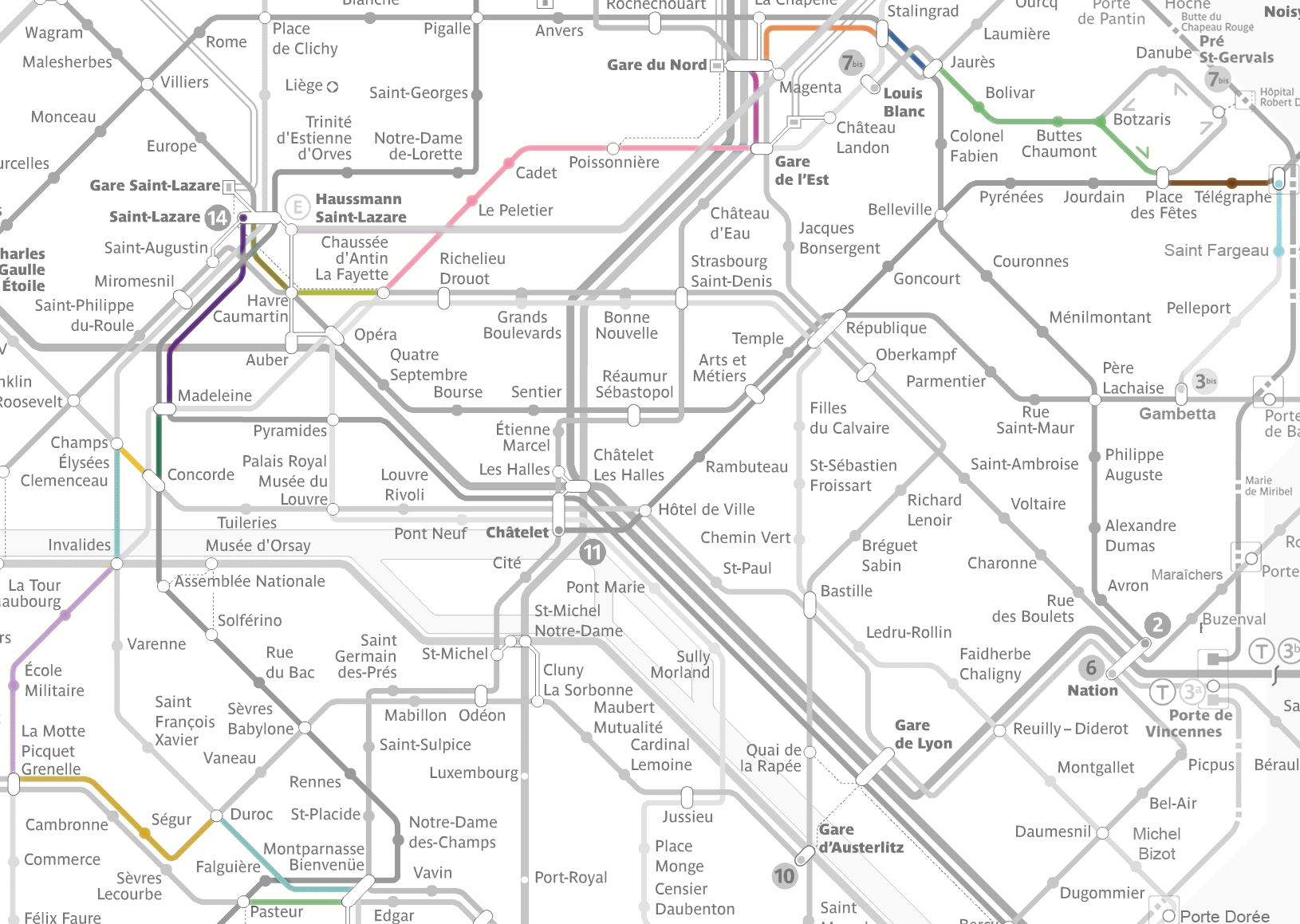}
\caption{Map for an optimal journey visiting all 16 metro lines without reusing the same station twice, from Pasteur to Saint Fargeau.}
\label{fig:metroNoRepet}
\end{figure}

\begin{figure}
\centering
\includegraphics[width=\linewidth]{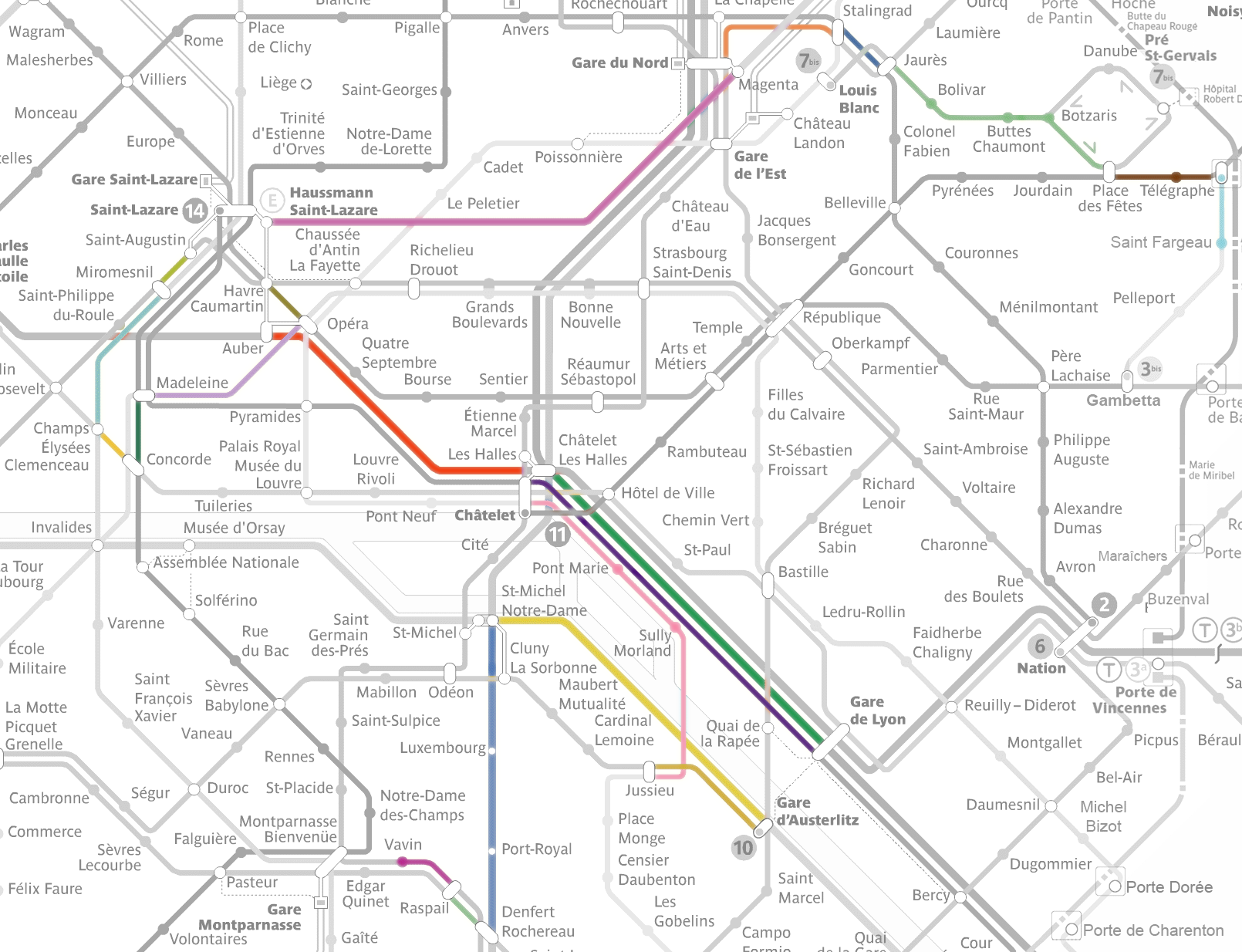}
\caption{Map for an optimal journey visiting all 16 metro lines and 5 RER lines without reusing the same station twice, from Vavin to Saint Fargeau.}
\label{fig:metroRERNoRepet}
\end{figure}

The details of a journey visiting all metro lines, from ``Pasteur'' to ``Saint-Fargeau'' is given in Table~\ref{table:metroNoRepet} and drawn in Figure~\ref{fig:metroNoRepet}.
The details of a journey visiting all metro lines and all RER lines, from ''Vavin'' to ``Saint-Fargeau'' is given in Table~\ref{table:metroRERNoRepet} and drawn in Figure~\ref{fig:metroRERNoRepet}.

One could also wonder if taking a line which was already used in the solution should be forbidden.
In this case, adding this constraint does not increase the number of mandatory steps, but there are now 8 ways to perform the metro journey and also 8 ways to perform the metro and RER journey.

\medskip

In how many steps can we visit all lines and also return to the starting point?
In other words, the solution must now be a cycle.
To solve this question, we can use the fact that any solution must use at least one endpoint of the line 3bis.
We remove Constraint~\ref{constr:degree_st} in the formulation and use an endpoint of the line 3bis instead of $s$ in Constraint~\ref{constr:flow1}.
It is worth noting that with a cyclic solution, one can start the journey at any station of the solution, which could be more convenient for the user.

\begin{funfact}
If one wants to be back at the starting point after the journey, it is possible to visit all 16 metro lines in 39 steps (allowing only metro lines).
\end{funfact}

\begin{figure}
\centering
\includegraphics[width=\linewidth]{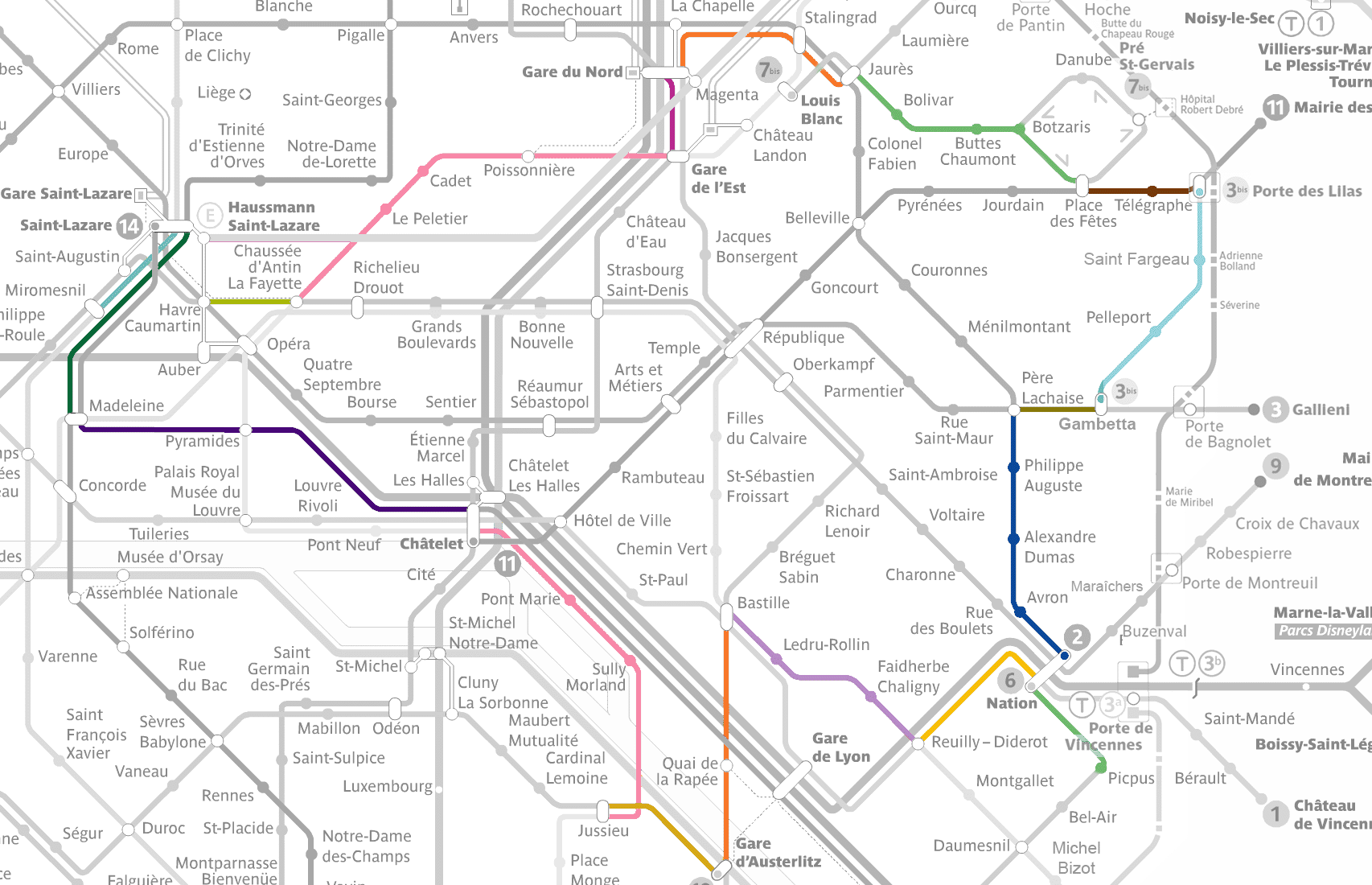}
\caption{Map for an optimal cyclic journey using all metro lines.}
\label{fig:metroCylic}
\end{figure}

One of the corresponding journey can be found in Table~\ref{table:metroCylic} and Figure~\ref{fig:metroCylic}. 
Note that under this setting, we reuse some lines and also some stations during the journey.
It would require 41 steps if reusing the same station is forbidden (and there are 132 ways to do it), 
and 42 steps if reusing the same line is also forbidden (and there are 2 ways to do it).

The next fact is quite surprising: it is \textit{faster} to visit \textit{more} lines!

\begin{funfact}
If one wants to be back at the starting point after the journey, it is possible to visit all 16 metro lines and all 5 RER lines in 33 steps, and there are 38 ways to do it.
\end{funfact}

\begin{figure}
\centering
\includegraphics[width=\linewidth]{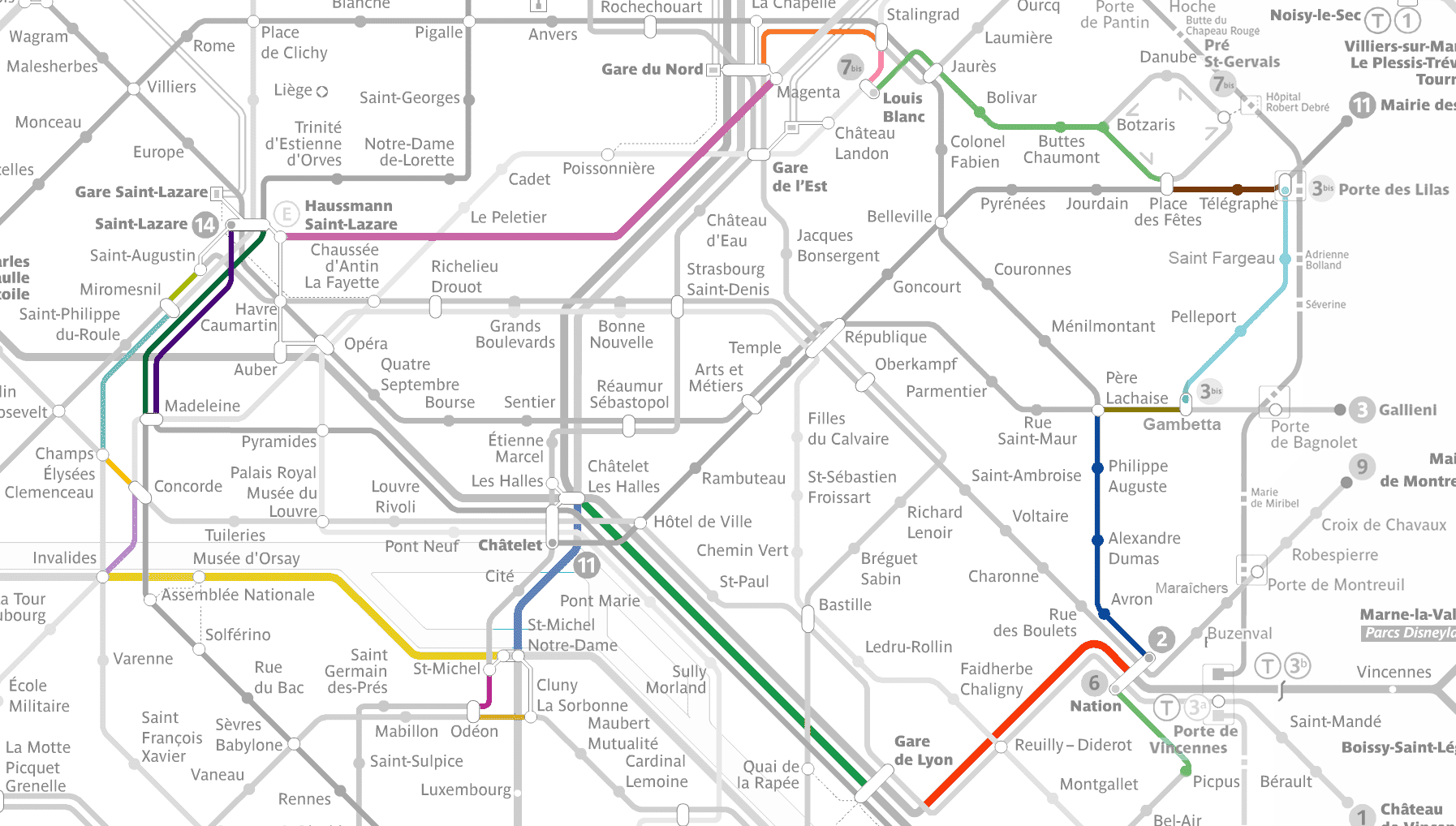}
\caption{Map for an optimal cyclic journey using all metro and RER lines.}
\label{fig:metroRERCylic}
\end{figure}

A possible journey can be found in Table~\ref{table:metroRERCylic} and Figure~\ref{fig:metroRERCylic}.
If reusing the same station is forbidden, then it is possible to do it in 38 steps (and there is only one way to do it). 

\subsection{Optimal journeys in Tokyo}

Motivated by a request from the researcher Takanori Maehara, the same program was applied to the network of Tokyo.
Two different companies are exploiting this metro network: ``Tokyo Metro'' and ``Toei''. 
The former manages 9 lines while the later manages 4 other lines. 
However, they are connected in some stations and one can use these lines without additional expense.
Therefore, we merged the data of these two networks into the same graph, and we added manually the missing connections.
The data for Tokyo Metro is taken from~\cite{Nakayama2015} and data for Toei is parsed from Wikipedia.

As the resulting network is quite dense, the journey is close to be optimal.

\begin{funfact}
It is possible to visit the 13 lines of Tokyo's  metro network with 15 steps, and there are 40 ways to do it.
\end{funfact}

\begin{figure}[ht!]
\centering

\includegraphics[width=\linewidth]{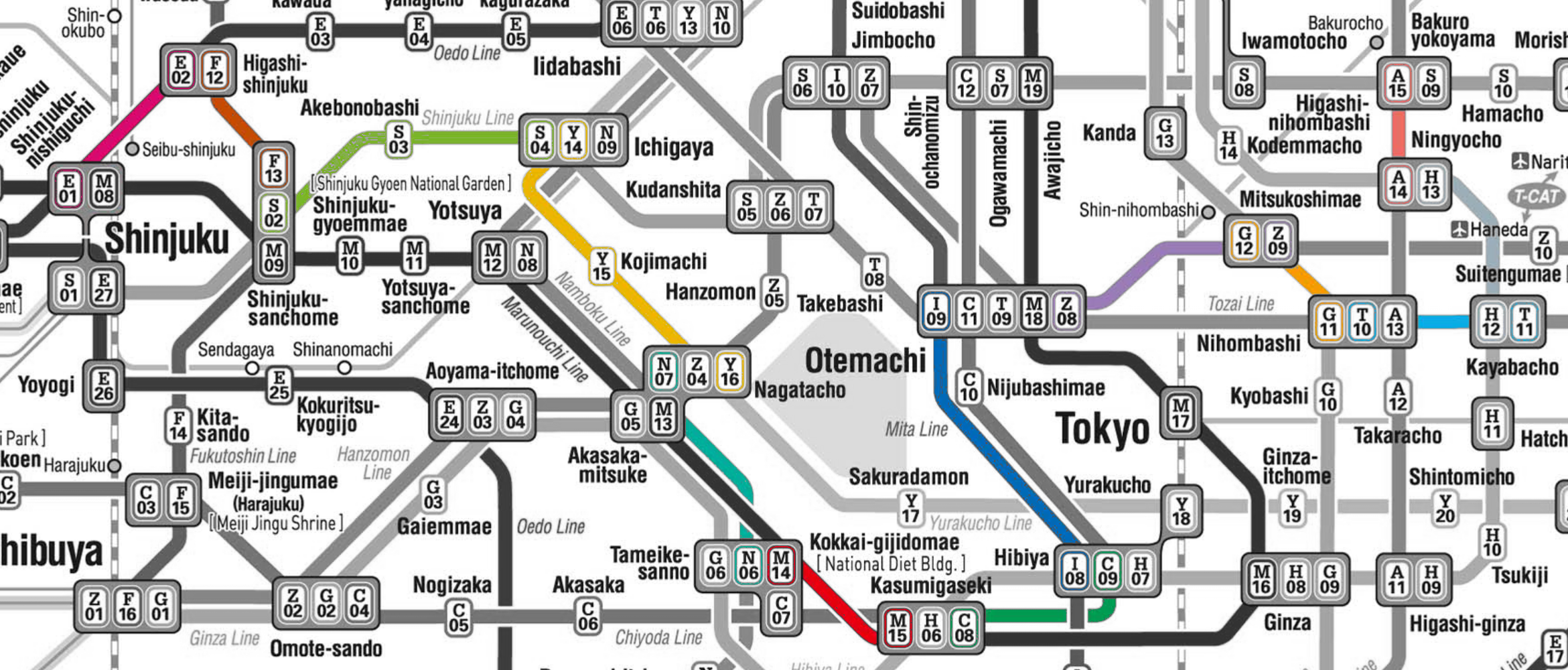}
\caption{Map for an optimal journey in Tokyo including all 13 metro lines in 15 steps, from Shinjuku-nishiguchi to Higashi-nihombashi.}
\label{fig:tokyo1}
\end{figure}

The details of a journey visiting all metro lines of Tokyo is given in Table~\ref{tab:tokyo1} and drawn in Figure~\ref{fig:tokyo1}.
This journey do not use multiple time the same station nor the same line, and allowing this do not decrease the size of the solution (but there are then 397 ways to do it!).

\begin{table}[ht!]
\centering
\resizebox{0.95\columnwidth}{!}{
\begin{tabular}{llll}
Step & Departure & Arrival & Line \\\hline
\rowcolor[RGB]{181, 30, 130}
1	&Shinjuku-nishiguchi&Higashi-shinjuku&Oedo Line\\
\rowcolor[RGB]{186, 104, 49}
2	&Higashi-shinjuku&Shinjuku-sanchome&Fukutoshin Line\\
\rowcolor[RGB]{109, 192, 103}
3   &Shinjuku-sanchome&Akebonobashi&Shinjuku Line\\
\rowcolor[RGB]{109, 192, 103}
4   &Akebonobashi&Ichigaya&Shinjuku Line\\
\rowcolor[RGB]{187, 139, 56}
5   &Ichigaya&Kojimachi&Yurakucho Line\\
\rowcolor[RGB]{187, 139, 56}
6   &Kojimachi&Nagatacho&Yurakucho Line\\
\rowcolor[RGB]{0, 181, 173}
7   &Nagatacho&Tameike-sanno&Namboku Line\\
\rowcolor[RGB]{237, 28, 36}
8   &Kokkai-gijido-mae&Kasumigaseki&Marunouchi Line\\
\rowcolor[RGB]{0, 166, 80}
9   &Kasumigaseki&Hibiya&Chiyoda Line\\
\rowcolor[RGB]{0, 128, 198}
10   &Hibiya&Otemachi&Mita Line\\
\rowcolor[RGB]{147, 124, 185}
11   &Otemachi&Mitsukoshimae&Hanzomon Line\\
\rowcolor[RGB]{247, 147, 29}
12   &Mitsukoshimae&Nihombashi&Ginza Line\\
\rowcolor[RGB]{0, 178, 221}
13   &Nihombashi&Kayabacho&Tozai Line\\
\rowcolor[RGB]{139, 162, 174}
14   &Kayabacho&Ningyocho&Hibiya Line\\
\rowcolor[RGB]{239, 91, 161}
15   &Ningyocho&Higashi-nihombashi&Asakusa Line
\end{tabular}
}
\caption{Description of the 15 steps to cover the 13 metro lines of Tokyo. Note that Tameike-sanno and Kokkai-gijido-mae are linked by a corridor.\label{tab:tokyo1}}
\end{table}

\subsection{In situ}

\begin{figure}[h!]
\begin{center}
\includegraphics[width=0.3\linewidth]{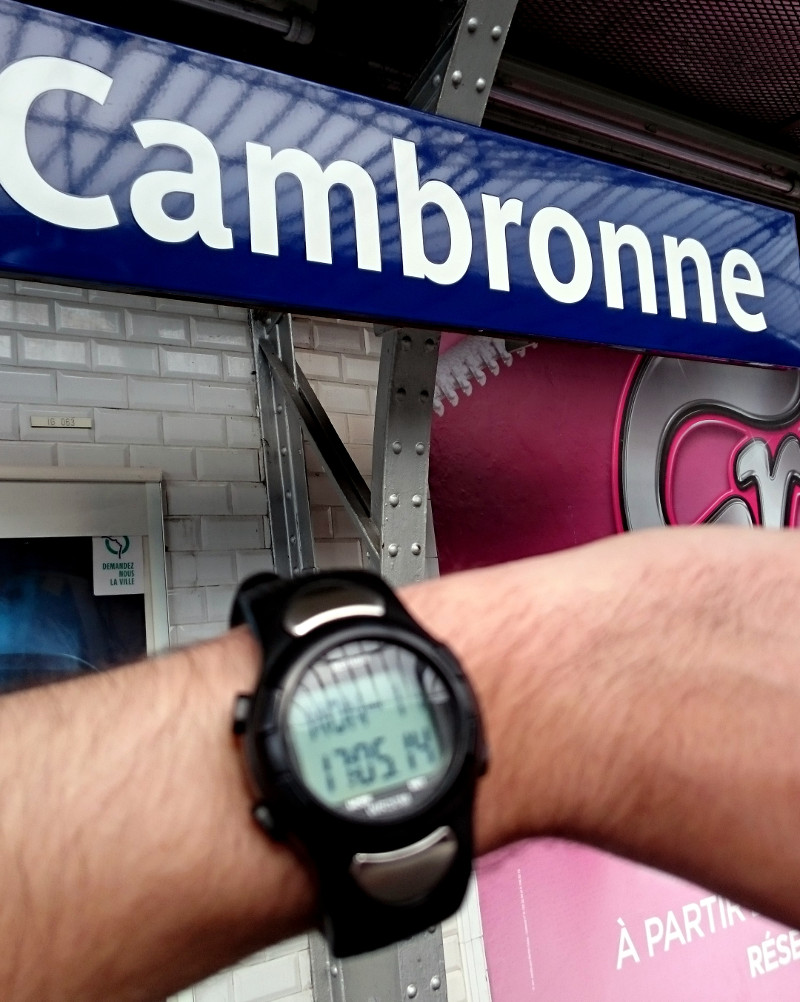}
\includegraphics[width=0.4\linewidth]{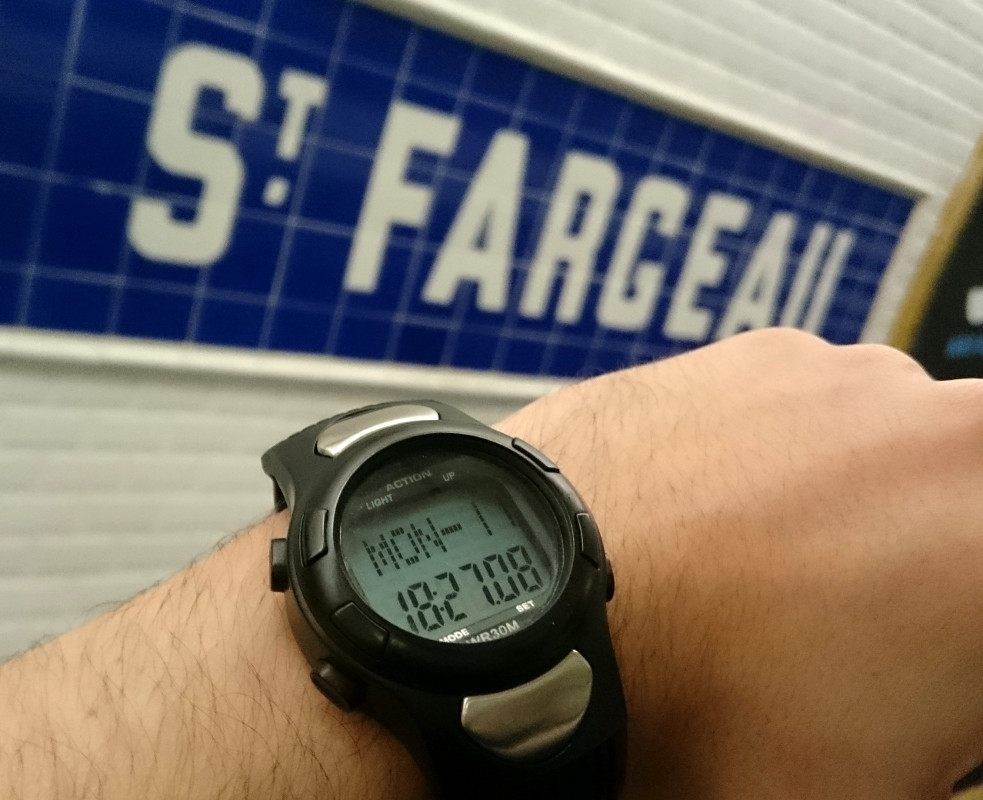}
\end{center}
\caption{Time at the starting point and the end point of the journey of Fun Fact~\ref{fact:metro}.\label{fig:montre}}
\end{figure}

On September 11, 2017, with my colleague Florian Yger, we did the journey of Fun Fact~\ref{fact:metro} in 1h22 (see Figure~\ref{fig:montre}).

Also, on December 19, 2017, Guillaume Py and Benoît Pecchio, who read a draft of this current paper, decided to do the journey of Fun Fact~\ref{fact:nostation} (the metro trip without reusing the same station twice). 
They needed 1h42 to perform this trip. 
They were very enthusiastic about this trip.

\section{Conclusion}

We presented a ``fun'' problem for traveling a metro network with an exotic constraint. 
We observed that this is a known problem in Operational Research and we proposed an ILP formulation. 
We also computed optimal journeys for different variations of the problem and tried one of them in practice.

We could probably be more accurate by taking into account distances between stations (an RER step is a bit more costly!), or taking into account the time to make a change of lines (you probably want to avoid Châtelet here).
It could be also possible to add a sort of ``bonus'' for passing by stations which are especially pretty or portion of lines with a nice view for an appropriate touristic walk.
Tourist Trip Design is actually an active field in Operational Research (see for example this survey~\cite{Gavalas2014}).
It is also possible to find a shortest walk or cycle starting from any specified station (This would give a solution which will be of length at least as big as the ones described here). 
This could be of interest for a user who would like to make the trip from a specified position. 
A website or a mobile application could be then derived, without any requirement of Internet access or heavy CPU usage with a pre-computation of all the possibilities.

\section*{Acknowledgments}

Thanks to Jérôme Lang for introducing this problem to me, to Florian Yger for taking the tour with me, and thanks to all my colleagues who had to suffer from my obsession with this problem.
Many thanks to Guillaume Py and Benoît Pecchio for having the trip and for letting me know about that.


\bibliographystyle{plain}
\bibliography{metro}

\begin{thebibliography}{10}

\bibitem{Avila2016}
Thais {\'{A}}vila, {\'{A}}ngel Corber{\'{a}}n, Isaac Plana, and
  Jos{\'{e}}~Mar{\'{\i}}a Sanchis.
\newblock {A New Branch-and-Cut Algorithm for the Generalized Directed Rural
  Postman Problem}.
\newblock {\em Transportation Science}, 50(2):750--761, 2016.

\bibitem{Cerny2015}
Pavol Cern{\'{y}}, Thomas~A. Henzinger, Laura Kov{\'{a}}cs, Arjun Radhakrishna,
  and Jakob Zwirchmayr.
\newblock {Segment Abstraction for Worst-Case Execution Time Analysis}.
\newblock In Jan Vitek, editor, {\em Proceedings of Programming Languages and
  Systems - 24th European Symposium on Programming, {ESOP} 2015, Held as Part
  of the European Joint Conferences on Theory and Practice of Software,
  {ETAPS}.}, volume 9032 of {\em LNCS}, pages 105--131. Springer, 2015.

\bibitem{Chassignet2004}
Philippe Chassignet.
\newblock
  \url{https://www.enseignement.polytechnique.fr/profs/informatique/Philippe.Chassignet/03-04/INF_431/td_1/lignes.data},
  2004.
\newblock [Online; accessed 03-July-2017 ].

\bibitem{Corberan2013}
{\'A}ngel Corber{\'a}n and Gilbert Laporte.
\newblock {\em Arc routing: problems, methods, and applications}.
\newblock SIAM, 2014.

\bibitem{Drexl2014}
Michael Drexl.
\newblock On the generalized directed rural postman problem.
\newblock {\em {JORS}}, 65(8):1143--1154, 2014.

\bibitem{Edmonds1973}
Jack Edmonds and Ellis~L. Johnson.
\newblock {Matching, Euler tours and the Chinese postman}.
\newblock {\em Math. Program.}, 5(1):88--124, 1973.

\bibitem{Gavalas2014}
Damianos Gavalas, Charalampos Konstantopoulos, Konstantinos Mastakas, and
  Grammati Pantziou.
\newblock A survey on algorithmic approaches for solving tourist trip design
  problems.
\newblock {\em Journal of Heuristics}, 20(3):291--328, Jun 2014.

\bibitem{GuinnessWorldRecords2015}
{Guinness World Records}.
\newblock {Fastest time to travel to all London Underground stations}.
\newblock
  \url{http://www.guinnessworldrecords.com/world-records/fastest-time-to-travel-to-all-london-underground-stations},
  2015.
\newblock [Online; accessed 20-February-2018].

\bibitem{GuinnessWorldRecords2017}
{Guinness World Records}.
\newblock {Fastest time to travel to all the Berlin U-Bahn metro stations}.
\newblock
  \url{http://www.guinnessworldrecords.com/world-records/fastest-time-to-travel-to-all-the-berlin-u-bahn-metro-stations},
  2017.
\newblock [Online; accessed 20-February-2018].

\bibitem{ILOG2012}
IBM ILOG.
\newblock {Inc. {CPLEX} 12.5 User's Manual}, 2012.

\bibitem{Tatou2015}
{Le Tatou}.
\newblock {Faire toutes les lignes du m\'etro en un seul jour ? LQC \#1.
  Youtube.}
\newblock \url{https://youtu.be/ZWjKbxCUCKc}, Jan 2015.
\newblock [Online; accessed 2-July-2017 ].

\bibitem{Lenstra1976}
Jan~Karel Lenstra and A.~H. G.~Rinnooy Kan.
\newblock On general routing problems.
\newblock {\em Networks}, 6(3):273--280, 1976.

\bibitem{Madore2004}
David Madore.
\newblock {Le D\'efi M\'etro}.
\newblock \url{http://www.madore.org/~david/misc/defimetro.html}, 2004.
\newblock [Online; accessed 2-July-2017 ].

\bibitem{Mole2011}
Annie Mole.
\newblock {Paris Subway Challenge by Adham Fisher. Going Underground's Blog}.
\newblock
  \url{http://london-underground.blogspot.fr/2011/10/paris-subway-challenge-by-adham-fisher.html},
  October 2011.
\newblock [Online; accessed 03-July-2017 ].

\bibitem{Nakayama2015}
Keitarou Nakayama.
\newblock {Tokyo Metro Stations}.
\newblock \url{http://linkdata.org/work/rdf1s3114i}, Apr 2015.
\newblock [Online; accessed 24-Jan-2018].

\bibitem{ANYSRC}
Peter~R. Samson.
\newblock {The Rise and Fall of the Amateur New York Subway Riding Committee}.
\newblock \url{http://www.gricer.com/anysrc/anysrc.html}, 1980.
\newblock [Online; accessed 27-June-2017 ].

\bibitem{Welz2014}
Wolfgang Welz.
\newblock {\em {Robot Tour Planning with High Determination Costs}}.
\newblock PhD thesis, TU Berlin, 2014.

\bibitem{WikipediaSubway}
Wikipedia.
\newblock {Subway Challenge --- Wikipedia{,} The Free Encyclopedia}.
\newblock
  \url{https://en.wikipedia.org/w/index.php?title=Subway_Challenge&oldid=816215959},
  2018.
\newblock [Online; accessed 20-February-2018].

\bibitem{HuffPost}
Alev Yildiz.
\newblock Il a parcouru toutes les stations du m\'etro parisien en un temps
  record {(Huffington Post)}.
\newblock
  \url{http://www.huffingtonpost.fr/2015/06/11/record-stations-metro-parisien-anglais-adham-fisher-transports-insolite_n_7561152.html},
  June 2015.
\newblock [Online; accessed 27-June-2017 ].

\end{thebibliography}


\begin{table}
\centering
\resizebox{0.9\columnwidth}{!}{
\begin{tabular}{llll}
Step & Departure & Arrival & Line \\\hline
\rowcolor[HTML]{6eca97}
1&Picpus&Nation&6	\\
\rowcolor[RGB]{209,48,47}
2&Nation&Gare de Lyon&A\\
\rowcolor[HTML]{5e9620}
3&Gare de Lyon&Chatelet - Les Halles&D\\
\rowcolor[HTML]{427bdb}
4&Chatelet - Les Halles&Saint-Michel - Notre-Dame&B  \\
\rowcolor[RGB]{207, 0, 158}
5 &Saint-Michel&Odeon&4   \\
\rowcolor[HTML]{c9910d}
6 &Odeon&Cluny - La Sorbonne&10 \\
\rowcolor[RGB]{252,217,70}
7&Saint-Michel - Notre-Dame&Musee d'Orsay&C \\
\rowcolor[RGB]{252,217,70}
8&Musee d'Orsay&Invalides&C \\
\rowcolor[RGB]{225, 155, 223}
9&Invalides&Concorde&8  \\
\rowcolor[RGB]{255, 205, 0}
10&Concorde&Champs Elysees - Clemenceau&1\\
\rowcolor[RGB]{ 	110, 196, 232}
11&Champs Elysees - Clemenceau&Miromesnil&13   \\
\rowcolor[HTML]{b6bd00}
12&Miromesnil&Saint-Augustin&9 \\
\rowcolor[HTML]{007852}
13&Saint-Lazare&Madeleine&12  \\
\rowcolor[HTML]{62259d}
\textcolor{white}{14}&\textcolor{white}{Madeleine}&\textcolor{white}{Pyramides}&\textcolor{white}{14} \\
\rowcolor[RGB]{ 	250, 154, 186}
15 &Pyramides&Opera&7  \\
\rowcolor[HTML]{837902}
16 & Opera&Havre - Caumartin&3  \\
\rowcolor[HTML]{bd76a1}
17 &Haussman St-Lazare&Magenta&E  \\
\rowcolor[RGB]{ 	255, 126, 46}
18 & Gare du Nord&Stalingrad&5 \\
\rowcolor[RGB]{0, 60, 166}
\textcolor{white}{19} & \textcolor{white}{Stalingrad}&\textcolor{white}{Jaures}&\textcolor{white}{2}   \\
\rowcolor[HTML]{6eca97}
20 &Jaures&Bolivar&7bis  \\
\rowcolor[HTML]{6eca97}
21 &Bolivar&Buttes Chaumont&7bis\\
\rowcolor[HTML]{6eca97}
22 &Buttes Chaumont&Botzaris&7bis  \\
\rowcolor[HTML]{6eca97}
23 &Botzaris&Place des Fetes&7bis \\
\rowcolor[HTML]{704b1c}
\textcolor{white}{24} & \textcolor{white}{Place des Fetes}&\textcolor{white}{Telegraphe}&\textcolor{white}{11} \\
\rowcolor[HTML]{704b1c}
\textcolor{white}{25} & \textcolor{white}{Telegraphe}&\textcolor{white}{Porte des Lilas}&\textcolor{white}{11}  \\
\rowcolor[HTML]{6ec4e8}
26 & Porte des Lilas&Saint-Fargeau&3bis 
\end{tabular} 
}
\caption{Description of the 26 steps to cover the 16 metro lines and the 5 RER lines. Note that according to the RATP map there is a corridor to join ``Saint-Michel - Notre-Dame'' and ``Saint-Michel'', ``Saint-Augustin'' and ``Saint-Lazare'', ``Havre-Caumartin'' and ``Haussman St-Lazare'', ``Magenta'' and ``Gare du Nord''.}
\label{table:rer}
\end{table}

\begin{table}
\centering
\begin{tabular}{llll}
Step & Departure & Arrival & Line \\\hline
\rowcolor[HTML]{6eca97}
1&Pasteur&Montparnasse - Bienvenue&6\\
\rowcolor[RGB]{ 	110, 196, 232}
2&Montparnasse - Bienvenue&Duroc&13\\
\rowcolor[HTML]{c9910d}
3&Duroc&Segur&10\\
\rowcolor[HTML]{c9910d}
4&Segur&La Motte-Picquet - Grenelle&10\\
\rowcolor[RGB]{225, 155, 223}
5&La Motte-Picquet - Grenelle&Ecole Militaire&8\\
\rowcolor[RGB]{225, 155, 223}
6&Ecole Militaire&La Tour-Maubourg&8\\
\rowcolor[RGB]{225, 155, 223}
7&La Tour-Maubourg&Invalides&8\\
\rowcolor[RGB]{ 	110, 196, 232}
8&Invalides&Champs Elysees - Clemenceau&13\\
\rowcolor[RGB]{255, 205, 0}
9&Champs Elysees - Clemenceau&Concorde&1\\
\rowcolor[HTML]{007852}
10&Concorde&Madeleine&12\\
\rowcolor[HTML]{62259d}
\textcolor{white}{11}&\textcolor{white}{Madeleine}&\textcolor{white}{Saint-Lazare}&\textcolor{white}{14}\\
\rowcolor[HTML]{837902}
12&Saint-Lazare&Havre - Caumartin&3\\
\rowcolor[HTML]{b6bd00}
13&Havre - Caumartin&Chaussee d'Antin - La Fayette&9\\
\rowcolor[RGB]{ 	250, 154, 186}
14&Chaussee d'Antin - La Fayette&Le Peletier&7\\
\rowcolor[RGB]{ 	250, 154, 186}
15&Le Peletier&Cadet&7\\
\rowcolor[RGB]{ 	250, 154, 186}
16&Cadet&Poissonniere&7\\
\rowcolor[RGB]{ 	250, 154, 186}
17&Poissonniere&Gare de l'Est&7\\
\rowcolor[RGB]{207, 0, 158}
18&Gare de l'Est&Gare du Nord&4\\
\rowcolor[RGB]{ 	255, 126, 46}
19&Gare du Nord&Stalingrad&5\\
\rowcolor[RGB]{0, 60, 166}
\textcolor{white}{20}&\textcolor{white}{Stalingrad}&\textcolor{white}{Jaures}&\textcolor{white}{2}\\
\rowcolor[HTML]{6eca97}
21&Jaures&Bolivar&7bis\\
\rowcolor[HTML]{6eca97}
22&Bolivar&Buttes Chaumont&7bis\\
\rowcolor[HTML]{6eca97}
23&Buttes Chaumont&Botzaris&7bis\\
\rowcolor[HTML]{6eca97}
24&Botzaris&Place des Fetes&7bis\\
\rowcolor[HTML]{704b1c}
\textcolor{white}{25}&\textcolor{white}{Place des Fetes}&\textcolor{white}{Telegraphe}&\textcolor{white}{11}\\
\rowcolor[HTML]{704b1c}
\textcolor{white}{26}&\textcolor{white}{Telegraphe}&\textcolor{white}{Porte des Lilas}&\textcolor{white}{11}\\
\rowcolor[HTML]{6ec4e8}
27&Porte des Lilas&Saint-Fargeau&3bis
\end{tabular}
\caption{An optimal journey with 27 steps visiting all 16 metro lines, without reusing the same station twice.}
\label{table:metroNoRepet}
\end{table}

\begin{table}
\centering
\begin{tabular}{llll}
Step & Departure & Arrival & Line \\\hline
\rowcolor[RGB]{207, 0, 158}
1&Vavin&Raspail&4   \\
\rowcolor[HTML]{6eca97}
2&Raspail&Denfert Rochereau&6   \\
\rowcolor[HTML]{427bdb}
3&Denfert Rochereau&Port Royal&B   \\
\rowcolor[HTML]{427bdb}
4&Port Royal&Luxembourg&B   \\
\rowcolor[HTML]{427bdb}
5&Luxembourg&Saint-Michel - Notre-Dame&B   \\
\rowcolor[RGB]{252,217,70}
6&Saint-Michel - Notre-Dame&Gare d'Austerlitz&C   \\
\rowcolor[HTML]{c9910d}
7&Gare d'Austerlitz&Jussieu&10   \\
\rowcolor[RGB]{ 	250, 154, 186}
8&Jussieu&Sully - Morland&7   \\
\rowcolor[RGB]{ 	250, 154, 186}
9&Sully - Morland&Pont Marie&7   \\
\rowcolor[RGB]{ 	250, 154, 186}
10&Pont Marie&Chatelet&7   \\
\rowcolor[HTML]{62259d}
\textcolor{white}{11}&\textcolor{white}{Chatelet}&\textcolor{white}{Gare de Lyon}&\textcolor{white}{14}   \\
\rowcolor[HTML]{5e9620}
12&Gare de Lyon&Chatelet - Les Halles&D   \\
\rowcolor[RGB]{209,48,47}
13&Chatelet - Les Halles&Auber&A   \\
\rowcolor[HTML]{837902}
14&Havre - Caumartin&Opera&3   \\
\rowcolor[HTML]{837902}
\rowcolor[RGB]{225, 155, 223}
15&Opera&Madeleine&8   \\
\rowcolor[HTML]{007852}
16&Madeleine&Concorde&12   \\
\rowcolor[RGB]{255, 205, 0}
17&Concorde&Champs Elysees - Clemenceau&1   \\
\rowcolor[RGB]{ 	110, 196, 232}
18&Champs Elysees - Clemenceau&Miromesnil&13   \\
\rowcolor[HTML]{b6bd00}
19&Miromesnil&Saint-Augustin&9   \\
\rowcolor[HTML]{bd76a1}
20&Haussman St-Lazare&Magenta&E   \\
\rowcolor[RGB]{ 	255, 126, 46}
21&Gare du Nord&Stalingrad&5   \\
\rowcolor[RGB]{0, 60, 166}
\textcolor{white}{22}&\textcolor{white}{Stalingrad}&\textcolor{white}{Jaures}&\textcolor{white}{2}   \\
\rowcolor[HTML]{6eca97}
23&Jaures&Bolivar&7bis   \\
\rowcolor[HTML]{6eca97}
24&Bolivar&Buttes Chaumont&7bis   \\
\rowcolor[HTML]{6eca97}
25&Buttes Chaumont&Botzaris&7bis   \\
\rowcolor[HTML]{6eca97}
26&Botzaris&Place des Fetes&7bis   \\
\rowcolor[HTML]{704b1c}
\textcolor{white}{27}&\textcolor{white}{Place des Fetes}&\textcolor{white}{Telegraphe}&\textcolor{white}{11}   \\
\rowcolor[HTML]{704b1c}
\textcolor{white}{28}&\textcolor{white}{Telegraphe}&\textcolor{white}{Porte des Lilas}&\textcolor{white}{11}   \\
\rowcolor[HTML]{6ec4e8}
29&Porte des Lilas&Saint-Fargeau&3bis
\end{tabular}
\caption{Optimal journey with 29 steps using all 16 metro lines and all 5 RER lines, without reusing the same station twice.}
\label{table:metroRERNoRepet}
\end{table}

\begin{table}
\centering
\begin{tabular}{llll}
Step & Departure & Arrival & Line \\\hline
\rowcolor[HTML]{837902}
1 & Gambetta & Pere Lachaise & 3   \\
\rowcolor[RGB]{0, 60, 166}
\textcolor{white}{2} & \textcolor{white}{Pere Lachaise} & \textcolor{white}{Philippe Auguste} & \textcolor{white}{2}   \\
\rowcolor[RGB]{0, 60, 166}
\textcolor{white}{3} &\textcolor{white}{Philippe Auguste }& \textcolor{white}{Alexandre Dumas} &\textcolor{white}{ 2}   \\
\rowcolor[RGB]{0, 60, 166}
\textcolor{white}{4}&\textcolor{white}{Alexandre Dumas} & \textcolor{white}{Avron} & \textcolor{white}{2}   \\
\rowcolor[RGB]{0, 60, 166}
\textcolor{white}{5}&\textcolor{white}{Avron} & \textcolor{white}{Nation} & \textcolor{white}{2}   \\
\rowcolor[HTML]{6eca97}
6   & Nation & Picpus & 6\\
\rowcolor[HTML]{6eca97}
7   & Picpus & Nation & 6\\
\rowcolor[RGB]{255, 205, 0}
8&Nation & Reuilly - Diderot & 1   \\
\rowcolor[RGB]{225, 155, 223}
9   & Reuilly - Diderot & Faidherbe - Chaligny & 8\\
\rowcolor[RGB]{225, 155, 223}
10   & Faidherbe - Chaligny & Ledru Rollin & 8\\
\rowcolor[RGB]{225, 155, 223}
11   & Ledru Rollin & Bastille & 8\\
\rowcolor[RGB]{ 	255, 126, 46}
12&Bastille & Quai de la Rapee & 5   \\
\rowcolor[RGB]{ 	255, 126, 46}
13&Quai de la Rapee & Gare d'Austerlitz & 5   \\
\rowcolor[HTML]{c9910d}
14&Gare d'Austerlitz & Jussieu & 10   \\
\rowcolor[RGB]{ 	250, 154, 186}
15&Jussieu & Sully - Morland & 7   \\
\rowcolor[RGB]{ 	250, 154, 186}
16&Sully - Morland & Pont Marie & 7   \\
\rowcolor[RGB]{ 	250, 154, 186}
17&Pont Marie & Chatelet & 7   \\
\rowcolor[HTML]{62259d}
\textcolor{white}{18}&\textcolor{white}{Chatelet} & \textcolor{white}{Pyramides} & \textcolor{white}{14}   \\
\rowcolor[HTML]{62259d}
\textcolor{white}{19}&\textcolor{white}{Pyramides} & \textcolor{white}{Madeleine} & \textcolor{white}{14}   \\
\rowcolor[HTML]{007852}
20&Madeleine & Saint-Lazare & 12   \\
\rowcolor[RGB]{ 	110, 196, 232}
21&Saint-Lazare & Miromesnil & 13   \\
\rowcolor[RGB]{ 	110, 196, 232}
22   & Miromesnil & Saint-Lazare & 13\\
\rowcolor[HTML]{b6bd00}
23&Havre - Caumartin & Chaussee d'Antin - La Fayette & 9   \\
\rowcolor[RGB]{ 	250, 154, 186}
24&Chaussee d'Antin - La Fayette & Le Peletier & 7   \\
\rowcolor[RGB]{ 	250, 154, 186}
25&Le Peletier & Cadet & 7   \\
\rowcolor[RGB]{ 	250, 154, 186}
26&Cadet & Poissonniere & 7   \\
\rowcolor[RGB]{ 	250, 154, 186}
27&Poissonniere & Gare de l'Est & 7   \\
\rowcolor[RGB]{207, 0, 158}
28&Gare de l'Est & Gare du Nord & 4   \\
\rowcolor[RGB]{ 	255, 126, 46}
29&Gare du Nord & Stalingrad & 5   \\
\rowcolor[RGB]{ 	255, 126, 46}
30&Stalingrad & Jaures & 5   \\
\rowcolor[HTML]{6eca97}
31&Jaures & Bolivar & 7bis   \\
\rowcolor[HTML]{6eca97}
32&Bolivar & Buttes Chaumont & 7bis   \\
\rowcolor[HTML]{6eca97}
33&Buttes Chaumont & Botzaris & 7bis   \\
\rowcolor[HTML]{6eca97}
34&Botzaris & Place des Fetes & 7bis   \\
\rowcolor[HTML]{704b1c}
\textcolor{white}{35}&\textcolor{white}{Place des Fetes} & \textcolor{white}{Telegraphe} & \textcolor{white}{11}   \\
\rowcolor[HTML]{704b1c}
\textcolor{white}{36}&\textcolor{white}{Telegraphe} & \textcolor{white}{Porte des Lilas} & \textcolor{white}{11}   \\
\rowcolor[HTML]{6ec4e8}
37&Porte des Lilas & Saint-Fargeau & 3bis   \\
\rowcolor[HTML]{6ec4e8}
38&Saint-Fargeau & Pelleport & 3bis   \\
\rowcolor[HTML]{6ec4e8}
39&Pelleport & Gambetta & 3bis   \\
\end{tabular}
\caption{One optimal cyclic journey with 39 steps using all 16 metro lines, allowing to use multiple time the same station.}
\label{table:metroCylic}
\end{table}

\begin{table}
\centering
\begin{tabular}{llll}
Step & Departure & Arrival & Line \\\hline
\rowcolor[HTML]{837902}
1&Gambetta & Pere Lachaise & 3   \\
\rowcolor[RGB]{0, 60, 166}
\textcolor{white}{2}&\textcolor{white}{Pere Lachaise} & \textcolor{white}{Philippe Auguste} &\textcolor{white}{ 2}   \\
\rowcolor[RGB]{0, 60, 166}
\textcolor{white}{3}&\textcolor{white}{Philippe Auguste }& \textcolor{white}{Alexandre Dumas} &\textcolor{white}{ 2}   \\
\rowcolor[RGB]{0, 60, 166}
\textcolor{white}{4}&\textcolor{white}{Alexandre Dumas} & \textcolor{white}{Avron} &\textcolor{white}{ 2}   \\
\rowcolor[RGB]{0, 60, 166}
\textcolor{white}{5}&\textcolor{white}{Avron} & \textcolor{white}{Nation} & \textcolor{white}{2}   \\
\rowcolor[HTML]{6eca97}
6 & Nation& Picpus& 6\\
\rowcolor[HTML]{6eca97}
7 & Picpus& Nation& 6\\
\rowcolor[RGB]{209,48,47}
8&Nation & Gare de Lyon & A   \\
\rowcolor[HTML]{5e9620}
9&Gare de Lyon & Chatelet - Les Halles & D   \\
\rowcolor[HTML]{427bdb}
10&Chatelet - Les Halles & Saint-Michel - Notre-Dame & B   \\
\rowcolor[RGB]{207, 0, 158}
11&Saint-Michel & Odeon & 4   \\
\rowcolor[HTML]{c9910d}
12   & Odeon& Cluny - La Sorbonne& 10\\
\rowcolor[RGB]{252,217,70}
13   & Saint-Michel - Notre-Dame &Musee d'Orsay& C\\
\rowcolor[RGB]{252,217,70}
14   &Musee d'Orsay &Invalides& C\\
\rowcolor[RGB]{225, 155, 223}
15&Invalides & Concorde & 8   \\
\rowcolor[RGB]{255, 205, 0}
16&Concorde & Champs Elysees - Clemenceau & 1   \\
\rowcolor[RGB]{ 	110, 196, 232}
17&Champs Elysees - Clemenceau & Miromesnil & 13   \\
\rowcolor[HTML]{b6bd00}
18&Miromesnil & Saint-Augustin & 9   \\
\rowcolor[HTML]{62259d}
\textcolor{white}{19} & \textcolor{white}{Saint-Lazare}& \textcolor{white}{Madeleine}& \textcolor{white}{14}\\
\rowcolor[HTML]{007852}
20 & Madeleine& Saint-Lazare& 12\\
\rowcolor[HTML]{bd76a1}
21&Haussman St-Lazare & Magenta & E   \\
\rowcolor[RGB]{ 	255, 126, 46}
22&Gare du Nord & Stalingrad & 5   \\
\rowcolor[RGB]{ 	250, 154, 186}
23   & Stalingrad& Louis Blanc& 7\\
\rowcolor[HTML]{6eca97}
24   & Louis Blanc& Jaures& 7bis\\
\rowcolor[HTML]{6eca97}
25&Jaures & Bolivar & 7bis   \\
\rowcolor[HTML]{6eca97}
26&Bolivar & Buttes Chaumont & 7bis   \\
\rowcolor[HTML]{6eca97}
27&Buttes Chaumont & Botzaris & 7bis   \\
\rowcolor[HTML]{6eca97}
28&Botzaris & Place des Fetes & 7bis   \\
\rowcolor[HTML]{704b1c}
\textcolor{white}{29}&\textcolor{white}{Place des Fetes} &\textcolor{white}{ Telegraphe }& \textcolor{white}{11}   \\
\rowcolor[HTML]{704b1c}
\textcolor{white}{30}&\textcolor{white}{Telegraphe} &\textcolor{white}{ Porte des Lilas} &\textcolor{white}{ 11}   \\
\rowcolor[HTML]{6ec4e8}
31&Porte des Lilas & Saint-Fargeau & 3bis   \\
\rowcolor[HTML]{6ec4e8}
32&Saint-Fargeau & Pelleport & 3bis   \\
\rowcolor[HTML]{6ec4e8}
33&Pelleport & Gambetta & 3bis   \\

\end{tabular}
\caption{An optimal cyclic journey with 33 steps using all 16 metro lines and all 5 RER lines, allowing to use the same station twice.}
\label{table:metroRERCylic}
\end{table}

\end{document}